# Evaluating the Robustness of Reinforcement Learning based Adaptive Traffic Signal Control

Dickens Kwesiga Ph.D.,[1] Angshuman Guin Ph.D.,[2] Khaled Abdelghany Ph.D.,[3] Michael Hunter Ph.D[4]

[1]Research Engineer, School of Civil and Environmental Engineering, Georgia Institute of Technology, Atlanta, USA, 30332; email: dkwesiga3@gatech.edu (Corresponding author)

[2]Principal Research Engineer, School of Civil and Environmental Engineering, Georgia Institute of Technology, Atlanta, USA, 30332; email: khaled@lyle.smu.edu

[3]Professor, Department of Civil and Environmental Engineering, Southern Methodist University, Dallas, USA, 75205; email: angshuman.guin@ce.gatech.edu

[4]Professor, School of Civil and Environmental Engineering, Georgia Institute of Technology, Atlanta, USA, 30332; email: michael.hunter@ce.gatech.edu

## ABSTRACT

Reinforcement learning (RL) has attracted increasing interest for adaptive traffic signal control due to its model-free ability to learn control policies directly from interaction with the traffic environment. However, several challenges remain before RL-based signal control can be considered ready for field deployment. Many existing studies rely on simplified signal timing structures, robustness of trained models under varying traffic demand conditions remains insufficiently evaluated, and runtime efficiency continues to pose challenges when training RL algorithms in traffic microscopic simulation environments. This study formulates an RL-based signal control algorithm capable of representing a full eight-phase ring–barrier configuration consistent with field signal controllers. The algorithm is trained and evaluated under varying traffic demand conditions and benchmarked against state-of-the-practice actuated signal control (ASC). To assess robustness, experiments are conducted across multiple traffic volumes and origin–destination (O–D) demand patterns with varying levels of structural similarity. To improve training efficiency, a distributed asynchronous training architecture is implemented that enables parallel simulation across multiple computing nodes. Results from a case study intersection show that the proposed RL-based signal control significantly outperforms optimized ASC, reducing average delay by 11–32% across movements. A model trained on a single O–D pattern generalizes well to similar unseen demand patterns but degrades under substantially different demand conditions. In contrast, a model trained on diverse O–D patterns demonstrates strong robustness, consistently outperforming ASC even under highly dissimilar unseen demand scenarios.

# 1      INTRODUCTION

Signal control strategies are commonly classified according to their responsiveness to traffic demand into fixed-time, actuated, and adaptive signal control systems. Fixed-time signal control operates using predetermined timing plans that allocate green time regardless of real-time traffic conditions. These plans are typically developed using historical traffic data and implemented through time-of-day (TOD) and day-of-week (DOW) schedules. Actuated signal control (ASC) improves upon fixed-time control by incorporating vehicle detection to extend or terminate phases based on real-time demand using rule-based logic. Adaptive signal control systems further extend this capability by continuously adjusting signal timing parameters based on observed or predicted traffic conditions. Widely deployed systems such as SCATS and SCOOT dynamically adjust cycle lengths, splits, and offsets across signal networks using traffic flow models and detector data.

Recent advances in machine learning (ML) have motivated the development of ML–based adaptive signal control systems, with most efforts focusing on reinforcement learning (RL). RL is particularly attractive for traffic signal control because modern RL algorithms are model-free and learn control policies directly from interaction with the environment, eliminating the need for explicit traffic prediction models. Numerous studies have demonstrated the potential of RL-based signal control to reduce vehicle delay, queue length, and travel time compared with conventional signal control strategies when evaluated in simulation environments. In many cases, RL-based approaches outperform fixed-time signal control and show promising results for adaptive traffic management.

Despite these encouraging results, several challenges remain before RL-based traffic signal control can be considered ready for real-world deployment. First, robustness of trained RL models under varying traffic demand conditions has received limited systematic evaluation. Many studies train and test RL models under identical or narrowly defined traffic demand scenarios, making it unclear how well the learned control policies generalize to different traffic patterns. Second, runtime efficiency remains a significant challenge when training RL models using microscopic traffic simulation, where thousands of simulation episodes and frequent interactions between the RL agent and the simulation environment can substantially increase computational requirements. Third, many existing RL formulations rely on simplified signal timing structures, such as limited phase sets or fixed phase sequences, which do not fully represent the complexity of field-deployed signal control systems based on ring–barrier configurations. Finally, RL-based signal control algorithms are often benchmarked primarily against fixed-time signal control, rather than ASC, which represents the prevailing state of practice in real-world signal operations.

To address these limitations, this study develops RL–based traffic signal control framework capable of representing a full eight-phase ring–barrier signal configuration consistent with field signal controllers. The proposed approach is trained and evaluated under varying traffic demand conditions and benchmarked against optimized ASC. To assess model robustness, extensive experiments are conducted using multiple origin–destination (O–D) demand patterns with varying levels of structural similarity. In addition, to address computational challenges associated with RL training in microscopic simulation, a parallelized asynchronous training architecture is implemented to improve runtime efficiency. The results demonstrate the effectiveness of the proposed approach and provide new insights into the robustness of RL-based traffic signal control under varying traffic demand conditions.

The remainder of the paper is organized as follows. The next section reviews relevant literature on RL–based traffic signal control and robustness evaluation. This is followed by a description of the proposed model formulation and the simulation framework used for training and evaluation. The subsequent section presents the experimental design and results of the case study analysis. Finally, conclusions and directions for future research are discussed.

## 2  RELATED STUDIES

### 2.1  RL for traffic signal control

A substantial body of recent work has focused on developing RL-based adaptive signal control systems (Bokade et al. 2023; Bouktif et al. 2023; Chang et al. 2024; Fu et al. 2023; Jia et al. 2024; Kolat et al. 2023; Li et al. 2025; Liu and Li 2023; Liu et al. 2023; Long and Chung 2023; Othman et al. 2025; Shen 2025; Shi et al. 2024; Shuvo et al. 2024; Wang et al. 2024; Xu et al. 2024). The appeal of RL-based signal control lies in the model free nature of state-of-the-art RL algorithms. Unlike the current field deployed adaptive signal control systems, model free RL-based adaptive signal control systems do not rely on state predictive models, making them more computationally efficient for real-time implementation.

Early research efforts largely focused on isolated intersection control, formulating single RL agents trained in microscopic traffic simulation to optimize signal timings at individual intersections (Aslani et al. 2019; Bálint et al. 2022; Bouktif et al. 2021; Casas 2017; Lee et al. 2022; Li et al. 2020; Li et al. 2016; Li et al. 2021; Liang et al. 2019; Liu et al. 2022; Liu et al. 2023; Shabestary et al. 2020; Wang et al. 2019). Signal coordination along arterials is central to traffic flow optimization. A key objective of coordinated signal systems is to promote cooperation between individual intersection controllers to progress traffic through adjacent intersections. Arterial signal timing optimization can be readily formulated as a fully cooperative multi agent reinforcement learning (MARL) problem with a single agent controlling each intersection and cooperating with the adjacent intersection agents to generate a coordinated signal timing plan. A growing number of efforts are extending single agent RL-based signal control to MARL-based signal control (Chang et al. 2024; Chen et al. 2021; Chu et al. 2019; Devailly et al. 2024; Fu et al. 2023; Jia et al. 2024; Kolat et al. 2023; Liu and Li 2023; Othman et al. 2025; Shen 2025; Shi et al. 2024; Shuvo et al. 2024; Wang et al. 2024; Wei et al. 2019).

RL-based signal control agents are typically developed and evaluated in traffic simulation environments including the commonly available off-the-shelf microscopic simulation packages such as SUMO (Bálint et al. 2022; Bao et al. 2023; Bokade et al. 2023; Fan and Yang 2024; Long and Chung 2023), PTV VISSIM® (Kwesiga et al. 2024), Aimsun (Casas 2017) and Paramics (Li et al. 2016). These platforms provide application programming interfaces (APIs), such as TraCI in SUMO and COM in PTV VISSIM®, which allow RL agents to interact with the simulation during runtime by observing traffic states and implementing signal control actions.

**R**untime efficiency remains a major challenge when training RL models using microscopic simulation**.** RL training commonly requires thousands of simulation episodes, each consisting of frequent state observations and control actions. These operations require repeated API communication between the RL agent and the simulator to retrieve traffic state information and update signal displays. As a result, simulation execution is frequently interrupted, significantly increasing computational overhead and slowing the training process. The challenge becomes more pronounced when high-resolution simulations or complex RL architectures are used. Consequently, improving runtime efficiency through efficient simulation environments, distributed execution, or parallel training architectures has become an important consideration in the development of scalable RL-based traffic signal control systems.

## 2.2 Robustness of RL-based Traffic signal control

A key challenge in deploying RL–based traffic signal control systems is ensuring robust performance under varying traffic demand conditions. In ML, robustness refers to the ability of a trained model to maintain stable performance when exposed to variations in input conditions. Traffic demand varies substantially across TOD, DOW, seasonal conditions, and unexpected disruptions. Conventional signal control strategies address this variability by developing multiple DOW and TOD plans optimized for different demand patterns. However, the robustness of RL-based signal control under changing traffic demand conditions has received limited attention in the literature. Many studies evaluate RL-based signal control using the same traffic demand conditions employed during training (Aslani et al. 2019; Bouktif et al. 2021; Liang et al. 2019; Wang et al. 2019). A smaller number of studies evaluate RL models under varying traffic volumes while maintaining fixed turning movement proportions (Shi et al. 2023).While these studies demonstrate improvements in metrics such as delay and queue length, they provide limited insight into how RL-based signal control performs when traffic demand patterns differ substantially from those used during training.

Only a few studies have examined robustness more explicitly. For example, Liu et al. (2022) evaluated RL models under modified traffic volumes while maintaining overall demand levels, and Lee et al. (2022) investigated model performance under alternative origin–destination (O–D) demand patterns. However, these efforts typically consider limited variations in demand or rely on simplified signal timing structures that do not reflect the complexity of field-deployed signal control systems. In addition, many RL studies benchmark performance only against fixed-time signal control rather than actuated signal control, which represents the current state of practice. Consequently, systematic evaluation of the robustness of RL-based traffic signal control under diverse traffic volumes and O–D demand patterns remains limited.

## 2.3 Literature summary

Despite the growing body of research on RL–based traffic signal control, several important challenges remain. First, the robustness of trained RL models under varying traffic demand conditions has received limited systematic evaluation, as many studies train and test models under identical or narrowly defined traffic scenarios. Second, runtime efficiency remains a practical challenge when training RL models in microscopic traffic simulation environments, where thousands of simulation episodes and frequent simulator–agent interactions significantly increase computational requirements. Third, many existing RL formulations rely on simplified signal timing structures that do not fully represent the complexity of field-deployed signal control systems, such as full ring–barrier configurations. Finally, RL-based signal control is frequently benchmarked against fixed-time signal control, rather than actuated signal control, which represents the prevailing state of practice in real-world operations. Addressing these limitations is essential for advancing RL-based signal control toward robust and deployable traffic management systems.

## 3 MODEL FORMULATION

## 3.1 Proximal policy optimization overview

This study adopts proximal policy optimization (PPO) as formulated by Schulman et al. (2017). PPO is a policy gradient (PG) algorithm that falls under the broad family of actor-critic algorithms. Accordingly, training PPO involves training both the actor and critic networks. The actor network learns the policy to select actions while the critic learns the value function to evaluate the performance of the actor network. PPO and its "older sibling" trust region policy optimization (TRPO) formulated by Schulman (2015) are aimed at improving the stability during training by limiting the change in policy for every step of gradient update.

The PPO objective function is based on importance sampling weights $\rho(s,a)$ which is the ratio of new policy $\pi_{new}$ to the old policy $\pi_{old}$ as in

$$\rho(s,a) = \frac{\pi_{new}(a|s;\varphi)}{\pi_{old}(a|s)}. \tag{1}$$

In Eq. (1), $\pi_{new}(a|s;\varphi)$ is the probability of the new policy selecting action, $a$ in state, $s$, $\varphi$ is the policy parameters while $\pi_{old}(a|s)$ is the probability of the old policy selecting action, $a$ in state, $s$.

The actor loss function, $\mathcal{L}(\varphi)$ is as in

$$\mathcal{L}(\varphi) = -\min\begin{pmatrix}\rho(s^t,a^t)Adv(s^t,a^t)\\ clip(\rho(s^t,a^t), 1-\epsilon, 1+\epsilon)Adv(s^t,a^t)\end{pmatrix} \tag{2}$$

$$where, \quad Adv(s,a) = \begin{cases} r^t & if\ s^{t+1}\ is\ terminal \\ r^t + \gamma V(s^{t+1}) - V(s^t) & otherwise \end{cases}$$

In Eq. (2), $\rho(s,a)$ is clipped with parameter $\epsilon$ to limit the change of the new policy from the old policy as discussed above. The term $Adv(s^t,a^t)$ is the action advantage, which comes from the critic where $V$ is the state value function and $r^t$ is the reward at time step $t$.

The critic loss function, $\mathcal{L}(\vartheta)$ is computes as in

$$\mathcal{L}(\vartheta) = (y^t - V(s^t;\vartheta)) \tag{3}$$

$$y^t = \begin{cases} r^t & if\ s^{t+1}\ is\ terminal \\ r^t + \gamma V^\pi(s^{t+1};\vartheta) & otherwise \end{cases}$$

In Eq. (3), $\mathcal{L}(\vartheta)$ is computed as the squared error of the state value estimate, $V^\pi$ and the bootstrapped target estimate $y^t$ where $\vartheta$ are the parameters of the critic network.

## 3.2 Definition of state, action and reward

### 3.2.1 State definition

The environment state at time step *t* denoted by $s^t$, is defined as a vector composed of two components: the traffic state and the signal state, as shown in Eq. (4).

$$state, \ s^t = [v_{11}, v_{12}, \ldots, v_{nm}, green\_dur_1, \ldots, green\_dur_{n_p}] \tag{4}$$

The first component represents the traffic state, defined as the number of vehicles *v* present in each lane of the intersection approaches. The resulting vector therefore has a dimension equal to the total number of approach lanes across the intersection. The formulation assumes that vehicle counts are available from connected vehicle data or video-based detection systems. The second component represents the signal state, which contains entries corresponding to the intersection signal phases ($n_p$). For phases currently receiving green, the corresponding entry records the elapsed green time ($green\_dur$), while all other entries are set to zero. This information is assumed to be available through real-time signal phase and timing (SPaT) data.

### 3.2.2 Action definition

In the proposed formulation, an action corresponds to selecting the next pair of compatible signal phases to activate. A phase is considered active when it is serving green time or clearance intervals (yellow and red clearance). Actions are defined based on a full dual ring–barrier configuration consisting of eight phases, as illustrated in Figure 1. The action space therefore consists of eight discrete actions, A = {1,2,3,4,5,6,7,8} where each action represents a pair of compatible phases, one from each ring. The feasible phase combinations are {1,5}, {1,6}, {2, 5}, {2, 6}, {3, 7}, {3, 8}, {4, 7} and {4,8}.

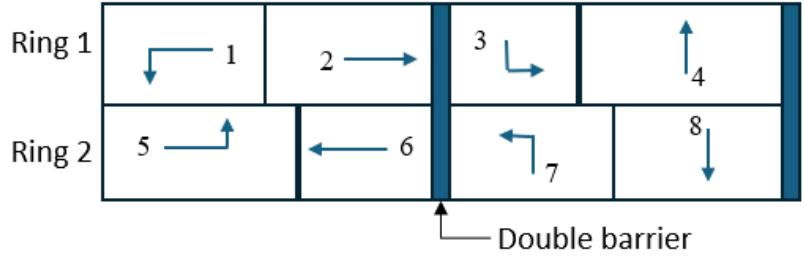

Figure 1. Action definition in a ring barrier diagram

The ring–barrier configuration ensures that conflicting movements are separated and that signal timing follows standard operational constraints. These constraints include satisfying minimum green times, clearance intervals, and barrier restrictions, which require both phases on one side of the barrier to terminate before phases on the opposite side can begin. Because of these constraints, not all actions are feasible at every decision point. At each decision point, a feasible action set is first determined based on the current signal state. The actor network produces a probability distribution over the full action space, and invalid action masking (IAM) algorithm formalized by Huang and Ontañón (2020) is used to prevent selection of infeasible actions by assigning large negative logits to invalid actions.

**Algorithm time step**

The RL algorithm operates using a decision interval $\Delta t$, which defines how long the selected action remains active before a new decision is made. This interval differs from the **s**imulation step, which determines how frequently the traffic simulation state is updated.

The decision interval is determined dynamically based on the state of each signal ring. For ring $r_j$, the ring-specific time step is defined as

$$\Delta tr_j \sim f(ring_{state}, ring_{tstate}, active_{phase}, committed_{phase}). \quad (5)$$

where the ring state indicates whether the ring is serving minimum green, extension green, or a clearance interval. The committed phase refers to the next phase that will be served after the current phase terminates. The overall algorithm time step is defined as the minimum of the two ring-specific time steps:

$$\Delta t = \min(\Delta tr_1, \Delta tr_2) \qquad (6)$$

This formulation ensures that signal timing constraints are satisfied while allowing the RL agent to make decisions only when phase transitions are feasible. Figure 2 illustrates an example of varying time steps for ring 1 and ring 2 at a given decision point.

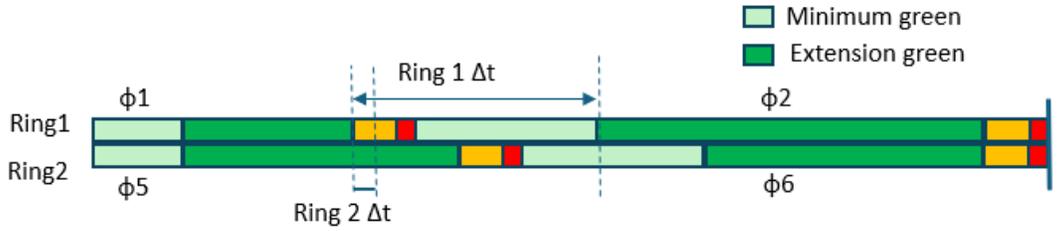

Figure 2. Determining algorithm timestep, Δt

### 3.2.3 Reward definition

The reward is defined as the negative sum of the normalized delay for all vehicles $V$ on all approaches of the intersection as in

$$r^t = -\sum_{v \in V} \frac{d_v^t}{d_{max}}. \qquad (7)$$

In Eq. (7), $d_v^t$ is the individual vehicle delay at time $t$ of each vehicle on the intersection approaches, while $d_{max}$ is the maximum expected vehicle delay for that intersection which can be estimated as a multiple of cycle length for a fixed time signal control. In the case study described in the next section, the algorithm is trained and tested in SUMO, a commonly used microscopic simulation environment in research. SUMO does not directly output vehicle delay but instead outputs "time loss" as a surrogate measure of delay. Delay is computed as the difference between actual travel time and free flow travel time.

## 4 CASE STUDY

### 4.1 Network Model

The formulated algorithm is trained for a hypothetical four-legged intersection. The main street (East-West) has two through lanes and a dedicated left turn lane in both directions. The cross street (North-South) has one through lane and one left turn lane in both directions. All left turns in the network have protected only phases. Right turn movements are omitted from the network without loss of generality. As this is a hypothetical network, default car following parameters in SUMO are left unchanged/uncalibrated.

### 4.2 RL-simulation architecture

Figure 3 shows SUMO and RL architecture during episode run and during model training. During each episode, the RL agent interacts with the SUMO simulation through the TraCI interface, which allows external programs to retrieve traffic state information and modify signal control parameters during runtime. At each decision point, the agent observes the environment state, determines the feasible action set based on ring–barrier constraints, and selects an action using the actor network. The selected signal displays are applied for a duration determined by the algorithm time step $\Delta t$. Resulting state transitions, rewards, actions, logits, and value estimates are stored in a memory buffer. At the end of each episode, the stored trajectories are used to update the actor and critic networks using the PPO objective functions defined in Eqs. (1)–(3).

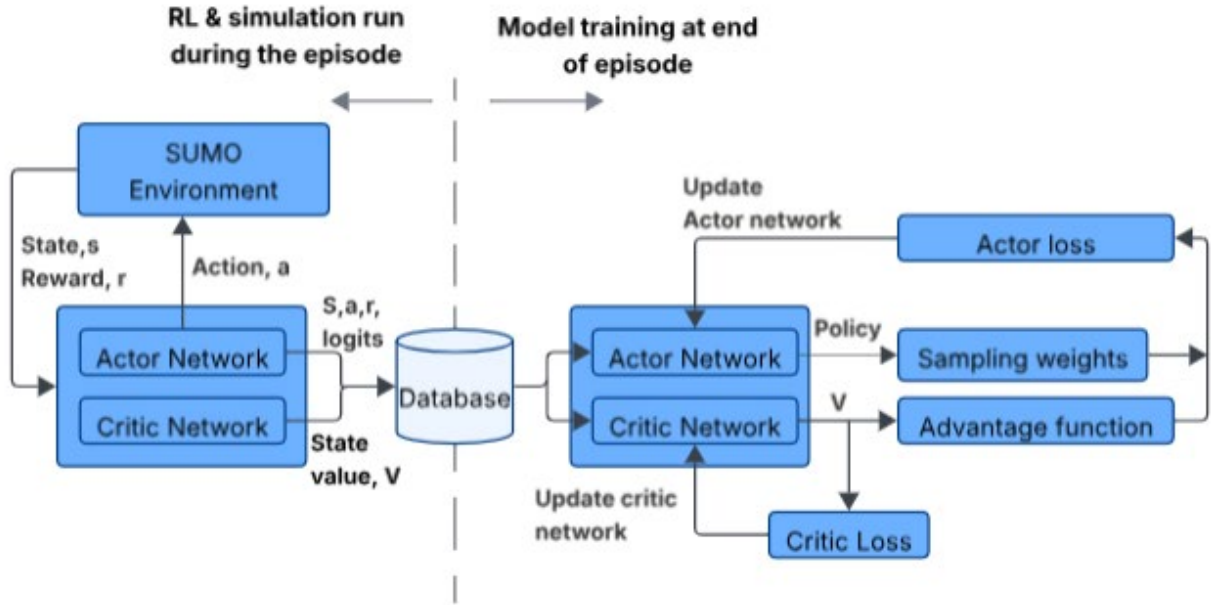

Figure 3. SUMO and RL architecture during episode run and during model training

Table 1 summarizes the final hyperparameters used for the actor and critic networks. Several combinations of hyperparameters were evaluated during preliminary experiments before selecting the final values reported here.

Table 1. Actor and critic network hyperparameters

| Hyperparameter | Actor network Value/s | Critic network Value/s |
|---|---|---|
| Hidden layer neurons | 64, 128, 256 | 64, 128, 256 |
| Clip ratio | 0.2 | |
| Learning rate | 0.0003 | 0.001 |
| Discount rate, gamma ($\gamma$) | | 0.99 |
| Entropy coefficient | 0.01 | |
| Maximum expected vehicle delay, $d_{max}$ | 300 seconds | |

### 4.2.1 Distributed training architecture

To address the computational demands of reinforcement learning (RL) training in microscopic traffic simulation, this study adopts a distributed and asynchronous training architecture that enables large-scale simulation experiments across multiple computing nodes. Training RL agents for traffic signal control typically requires thousands of simulation episodes. In a conventional single-machine architecture, simulation and learning occur sequentially, resulting in long training times due to the frequent interaction between the RL agent and the simulator to retrieve traffic states and apply control actions. The proposed distributed architecture improves training efficiency by executing multiple simulation environments in parallel across several worker machines.

As illustrated in Figure 4, the architecture consists of two primary components: a central learner node and multiple worker nodes distributed across several PCs. Each worker machine runs multiple independent instances of the SUMO. These simulation instances interact with a local copy of the RL policy to generate environment trajectories consisting of states, actions, rewards, and value estimates. Experiences generated by each simulation instance are stored temporarily in local buffers on the worker machines and periodically transferred to a shared experience repository accessible by the learner node.

The learner PC aggregates experiences collected from all worker nodes and performs updates to the actor and critic networks. After each training update, the updated policy parameters are distributed back to the worker nodes, which continue generating new simulation experiences using the updated policy. Because simulation environments operate independently, decision times across environments do not need to be synchronized, enabling asynchronous experience generation and improved computational efficiency.

This architecture effectively decouples simulation from policy learning and allows training throughput to scale with the number of available computing resources. Additional worker machines can be incorporated with minimal modification to the training pipeline, enabling efficient large-scale training across many traffic demand scenarios and network configurations. By leveraging distributed simulation and asynchronous experience aggregation, the proposed framework significantly reduces RL training time while maintaining stable policy learning.

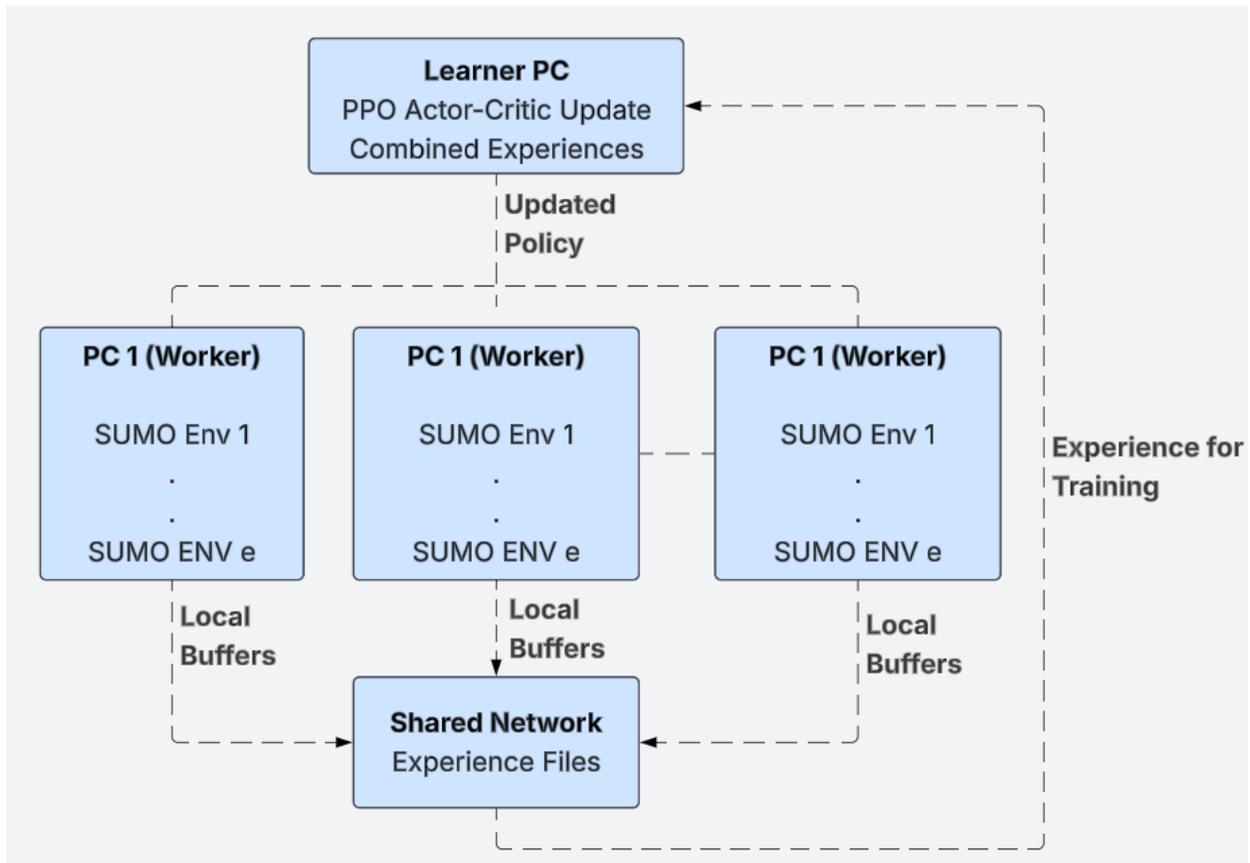

Figure 4. Distributed architecture for SUMO and RL

### 4.3    Experiment design

Experiments are designed to evaluate the performance and robustness of RL-based signal control under varying traffic volumes and origin–destination (O–D) patterns. Figure 5 illustrates the O–D patterns used in the training and testing experiments. The numbers shown represent traffic volumes in vehicles per hour, and the patterns are labeled A–F for reference in subsequent sections.

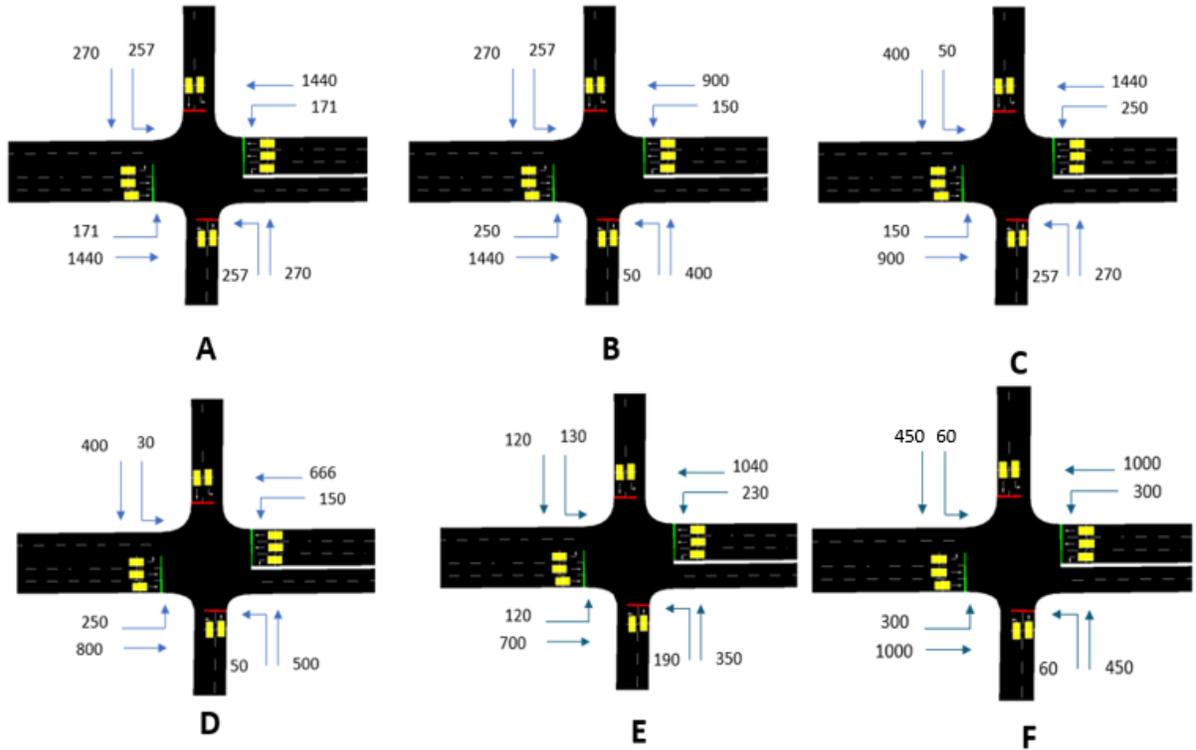

Figure 5. O-D patterns for training and testing

To quantify the similarity between O–D patterns, the structural similarity index (SSIM) is used. SSIM, originally proposed by Wang et al. (2004), measures the similarity between two matrices based on structural information and takes values between 0 and 1, with 1 indicating identical patterns. Following the formulation proposed by Djukic (2013), SSIM is applied to compare O–D matrices in this study. Table 2 presents the dissimilarity matrix (1 − SSIM) for the O–D patterns considered. Larger values indicate greater differences between traffic demand patterns, with the highest dissimilarity (0.603) observed between patterns A and D.

All O–D patterns are selected such that the overall intersection volume-to-capacity ratio (v/c) lies between 0.5 and 1.05. At low demand levels (v/c < 0.5), minimum green times dominate signal operation and limit the effectiveness of adaptive control strategies. Conversely, when demand substantially exceeds capacity (v/c > 1.0), both RL-based and actuated signal control tend to operate similarly to fixed-time control as most phases reach their maximum green times. Constraining the v/c range ensures that meaningful performance differences between control strategies can be observed.

Table 2. Dissimilarity (1-SSIM) of O-D patterns

| OD Pattern | OD Pattern | | | | | | | |
|---|---|---|---|---|---|---|---|---|
| | A | B | C | D | E | F | G | H |
| A | 0.000 | 0.198 | 0.198 | 0.603 | 0.216 | 0.455 | 0.378 | 0.309 |
| B | 0.198 | 0.000 | 0.479 | 0.349 | 0.394 | 0.347 | 0.321 | 0.322 |
| C | 0.198 | 0.479 | 0.000 | 0.491 | 0.205 | 0.347 | 0.284 | 0.228 |
| D | 0.603 | 0.349 | 0.491 | 0.000 | 0.536 | 0.103 | 0.131 | 0.185 |
| E | 0.216 | 0.394 | 0.205 | 0.536 | 0.000 | 0.438 | 0.361 | 0.307 |
| F | 0.455 | 0.347 | 0.347 | 0.103 | 0.438 | 0.000 | 0.034 | 0.087 |

### 4.3.1 Training and testing O-D patterns

Two RL models are trained using different traffic demand configurations. Model I is trained using a single O–D pattern (A), while Model II is trained using multiple O–D patterns (A, B, and C). Training is conducted using the distributed architecture described in Section 4.2, where multiple worker machines run independent SUMO simulation environments in parallel. Each worker executes one or more simulation instances that interact with the RL policy to generate experience trajectories.

For Model I, all environments are supplied with O–D pattern A, with stochastic vehicle arrivals introducing minor variations between simulation runs. For Model II, the environments are distributed across the worker machines such that different O–D patterns are used concurrently during training (A, B, and C). Experiences generated by all environments are aggregated by the learner node to update the policy parameters. O–D patterns D, E, and F are reserved exclusively for evaluating the trained models.

### 4.3.2 Actuated signal control baseline

The RL-based signal control is benchmarked against ASC, representing current practice. For each test O–D pattern, signal timings are optimized using Synchro. All movements are detected, resulting in a fully actuated control configuration, and the same minimum green and clearance intervals used in the RL formulation are applied. The optimized signal timings are implemented in SUMO to simulate the intersection under the ASC baseline. Figure 6 shows an example of signal timing optimization in Synchro.

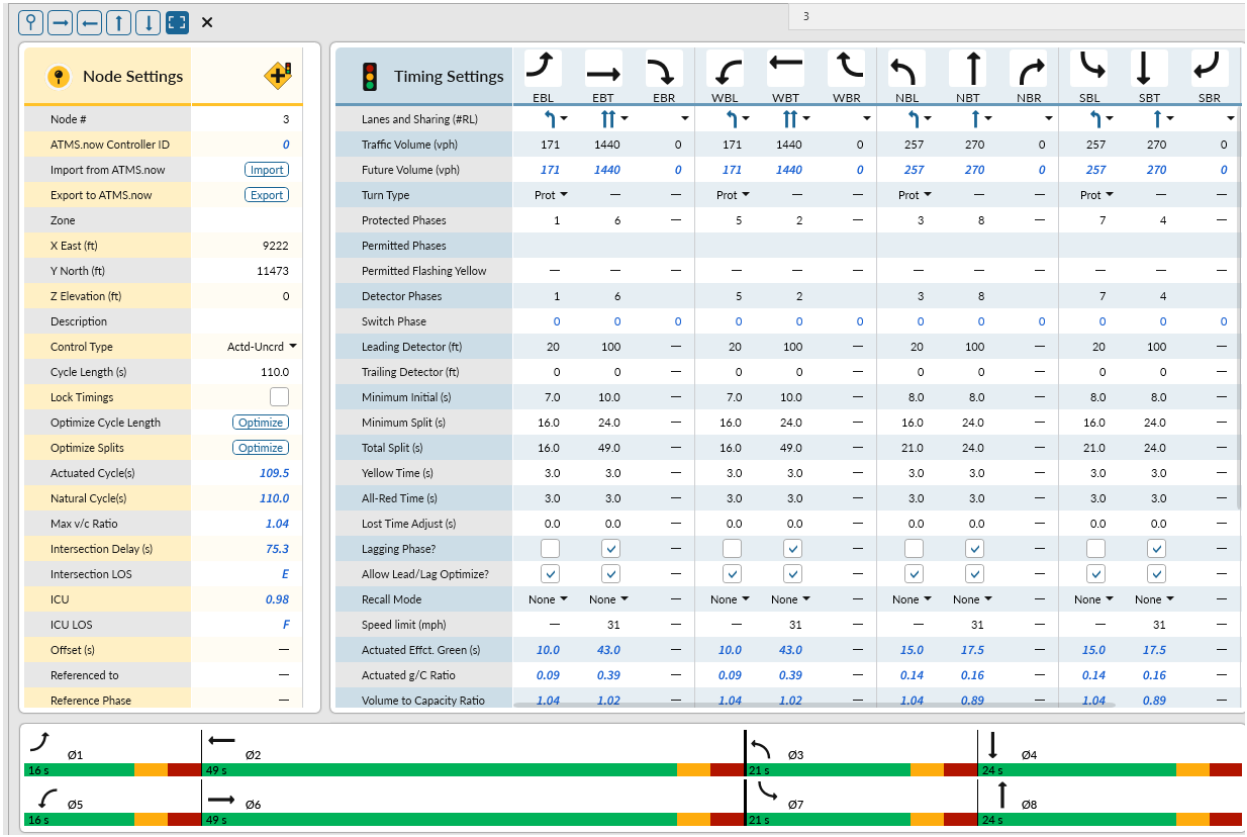

Figure 6. Signal optimization in Synchro

## 5 RESULTS AND ANALYSIS

### 5.1 Model training

Figure 7 shows the learning curves for the two RL models. The x-axis represents the training episode number (each episode corresponds to 3600 s of simulated traffic), while the y-axis shows the average reward across all decision steps within the episode. Model I converged more rapidly, stabilizing before approximately 500 episodes, whereas Model II reached a similar level of stability after about 1250 episodes. The slower convergence of Model II reflects the greater variability in traffic demand introduced by training on multiple O–D patterns. Although only the first 2000 episodes are shown in Figure 7, training continued to 5000 episodes, after which no significant improvement in reward values was observed.

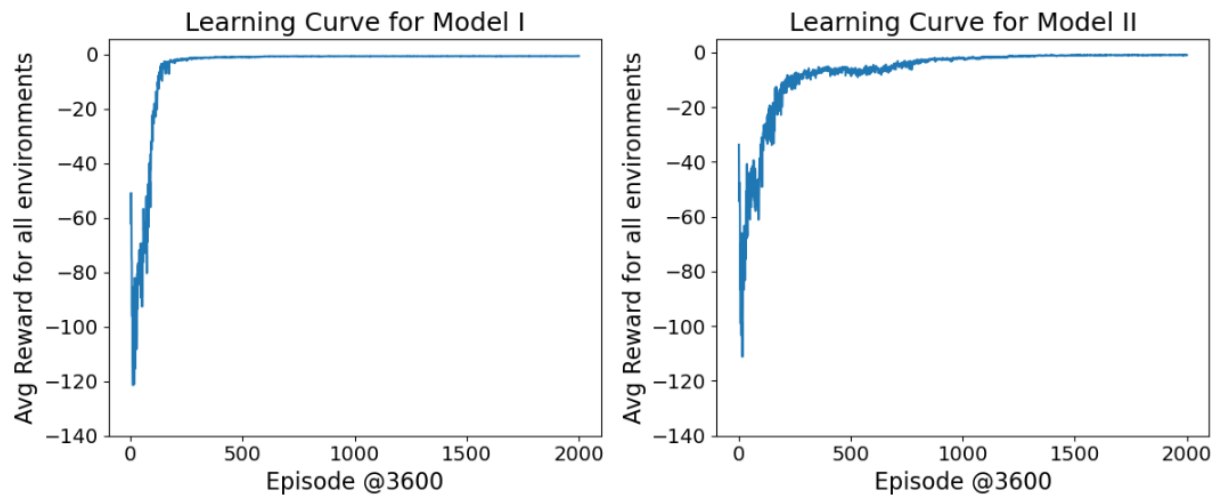

Figure 7. Training curves for the first 2000 episodes

## 5.2 Testing on O-D pattern already seen in training

Both RL models (Model I and Model II) are evaluated using O–D pattern A, which was included in the training data. ASC optimized for pattern A serves as the baseline. For each control strategy, 10 replicate simulation runs are performed, each lasting 2 hours. The first 15 min are used as a warm-up period, and performance measures are collected over the remaining 105 min.

Figure 8 presents the delay distributions for the eight intersection movements. Both RL models consistently outperform ASC across all movements. Table 3 summarizes the percentage reductions in average delay relative to ASC. The RL-based control achieves delay reductions ranging from 11.1% to 31.7% depending on the movement. The performance of the two RL models is largely comparable.

Table 3. Comparing the performance of RL-based SC against ASC for test O-D pattern A

| Movement | Model I<br>% change vs ASC | Model II<br>% change vs ASC |
|---|---|---|
| Eastbound Left (EBL) | 27.3 | 25.8 |
| Eastbound Through (EBT) | 11.1 | 11.1 |
| Westbound Left (WBL) | 23.1 | 18.5 |
| Westbound Through (WBT) | 13.6 | 11.4 |
| Northbound Left (NBL) | 29.6 | 29.6 |
| Northbound Through (NBT) | 30.0 | 31.7 |
| Southbound Left (SBL) | 29.7 | 23.4 |
| Southbound Through (SBT) | 28.3 | 30.0 |

Model II shows slightly higher delay for westbound left (WBL) and southbound left (SBL) movements, while delays for the remaining movements are nearly identical between the two models. Overall, the results indicate that training with multiple O–D patterns does not reduce performance when evaluated on a demand pattern already included in training.

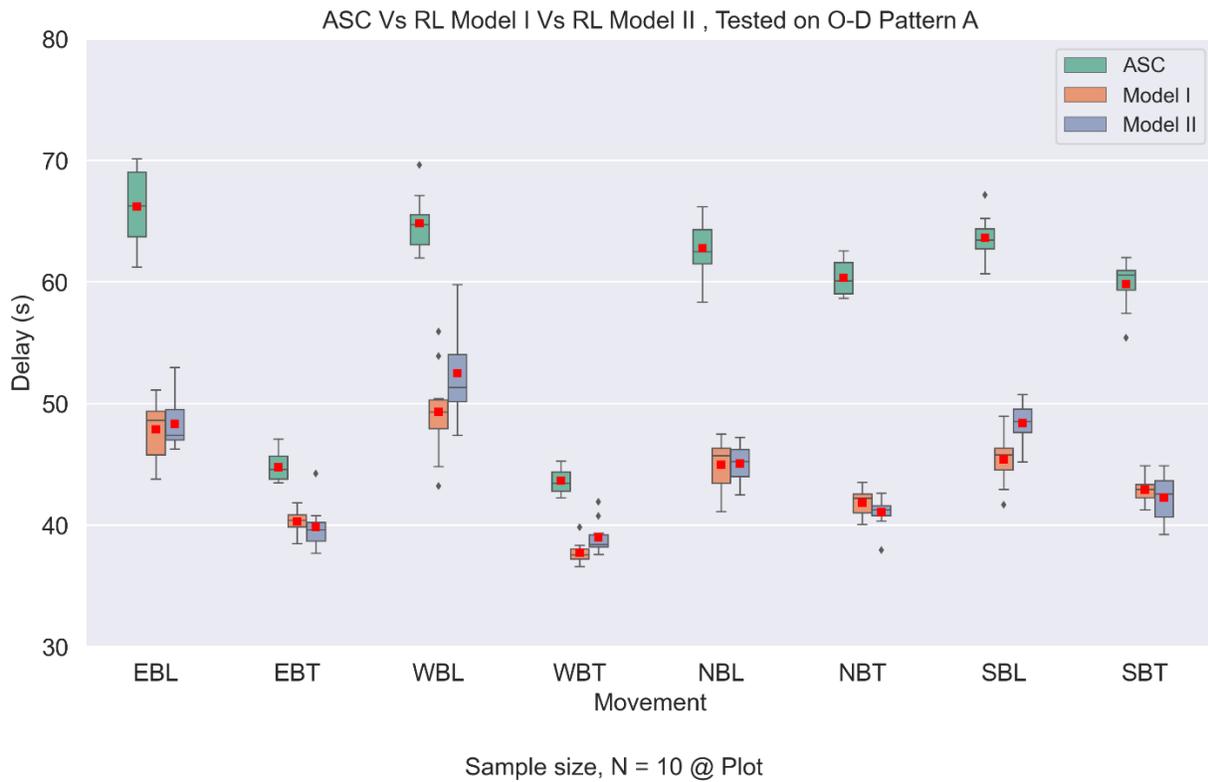

Figure 8. The performance of Model I and Model II when tested on O-D Pattern A compared to ASC.

## 5.3 Testing on O-D patterns unseen in the training

The trained RL models are next evaluated using O–D patterns not included in the training data. These test patterns are selected to be structurally dissimilar from the training patterns based on the SSIM measure described in Section 4.3. Figure 9 shows the performance of the two models for O–D pattern D, which exhibits the largest dissimilarity relative to the training patterns. The ASC baseline is optimized specifically for pattern D. Model II consistently outperforms ASC for most movements, particularly on the higher-volume approaches such as EBL, EBT, and WBT. Slightly higher delays occur for NBL and SBL, which have relatively low volumes (60 veh/h). Because the RL formulation minimizes overall intersection delay, movements with very small demand may receive less priority.

In contrast, Model I exhibits reduced performance under the unseen demand pattern, particularly for the EBL and EBT movements where delays exceed those of ASC. This behavior reflects the limited generalization capability of a model trained on a single O–D pattern. When the demand distribution changes substantially—particularly in the eastbound direction—Model I fails to adapt the learned phase sequence and timing effectively. These results highlight the improved robustness of Model II, which was trained using multiple demand patterns.

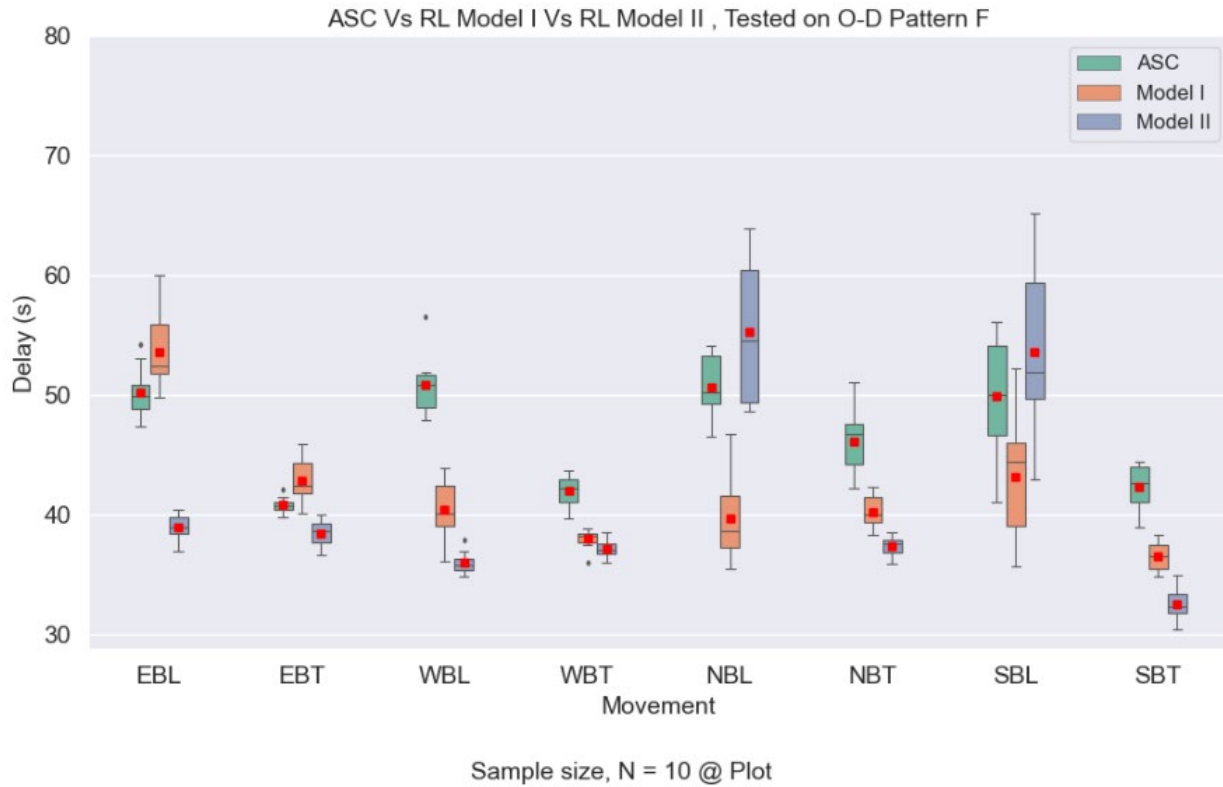

Figure 9. The performance of Model I and Model II when tested on O-D Pattern D compared to ASC.

Figure 10 presents the same comparison for O–D pattern E. Both RL models significantly outperform ASC across most movements, with the exception of WBT. The performance of Model I and Model II is largely comparable, except for WBL, where small differences are observed. O–D pattern E is more similar to the training pattern A (1 − SSIM = 0.216) than pattern D (1 − SSIM = 0.603) (Table 2). Consequently, Model I—trained solely on pattern A—performs substantially better on pattern E than on pattern D (Figure 9).

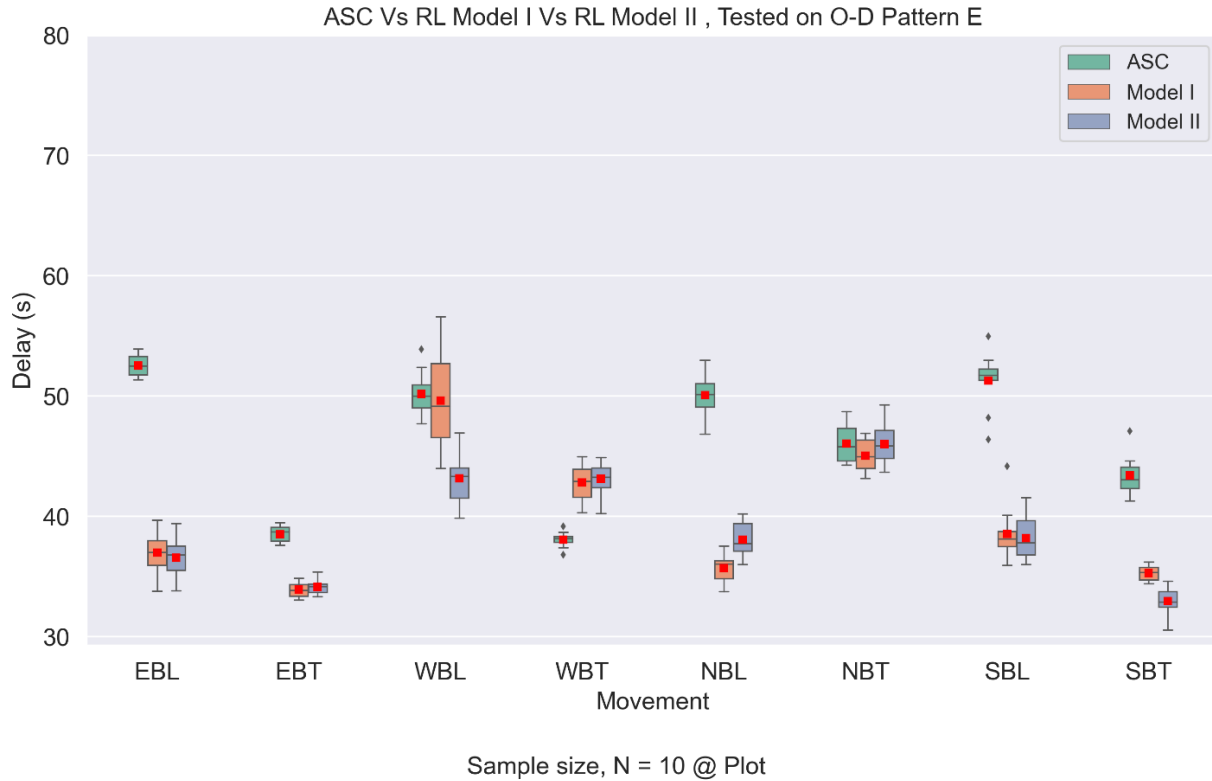

Figure 10. The performance of Model I and Model II when tested on O-D Pattern E compared to ASC.

Figure 11 shows the performance of the two models for O–D pattern F. For the higher-volume movements, particularly the main street left turns (EBL and WBL) and the side street through movements (NBT and SBT), Model II performs substantially better than Model I.. For NBL and SBL, Model II appears to slightly underperform relative to Model I. However, these movements have relatively small traffic volumes, and because the RL formulation minimizes overall intersection delay, movements with very low demand may receive less priority. Overall, the results for O–D pattern F are consistent with the earlier findings for patterns D and E. The model trained on multiple O–D patterns (Model II) demonstrates stronger robustness when applied to unseen demand conditions, particularly for higher-volume movements that dominate intersection delay, while the model trained on a single O–D pattern (Model I) exhibits more variable performance.

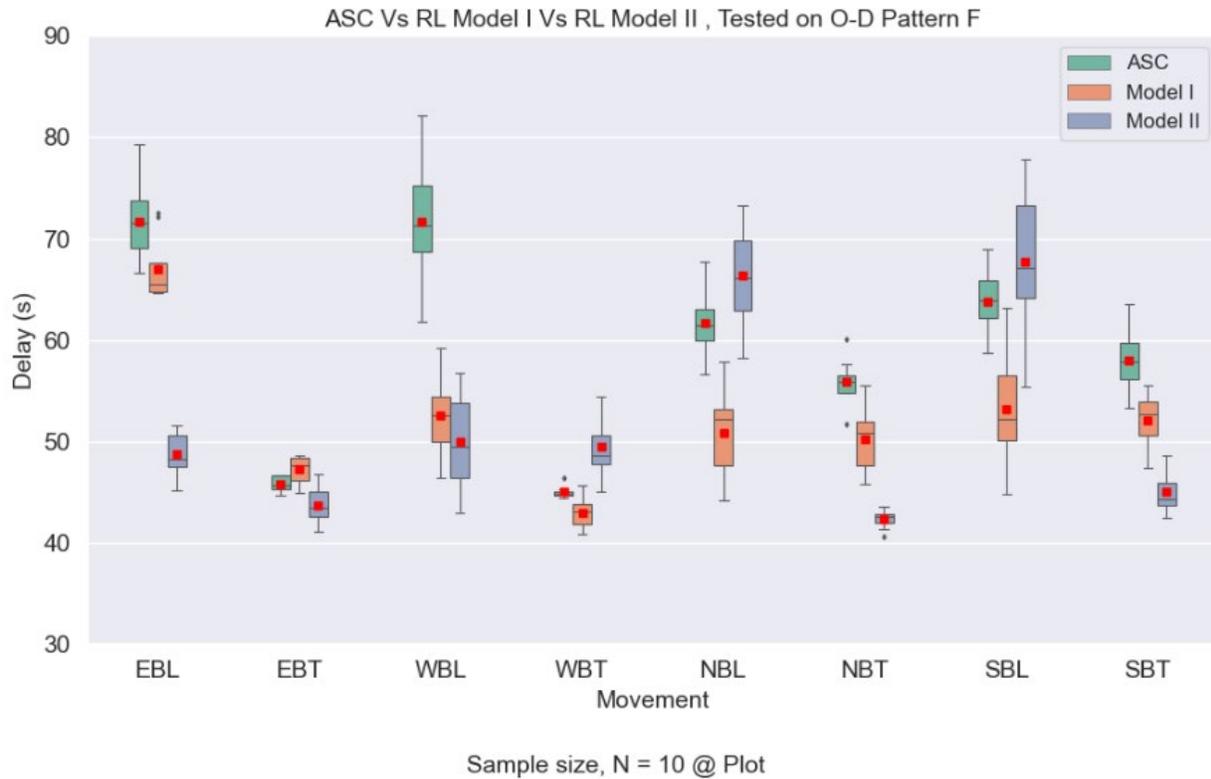

Figure 11. The performance of Model I and Model II when tested on O-D Pattern F compared to ASC.

## 6	CONCLUSION

This study develops RL–based traffic signal control algorithm capable of operating within a full eight-phase ring–barrier configuration representative of field signal control systems. The proposed approach is evaluated using a case study intersection under varying traffic volumes and origin–destination (O–D) demand patterns and benchmarked against optimized actuated signal control (ASC).

The results show that RL-based signal control can significantly outperform optimized ASC, with delay reductions ranging from 11% to 31% across different traffic movements. A model trained on a single O–D pattern generalizes well to demand patterns that are similar to the training conditions but exhibits reduced performance when applied to substantially different traffic distributions. In contrast, a model trained on multiple O–D patterns demonstrates stronger robustness, consistently outperforming ASC even under highly dissimilar unseen demand conditions.

These findings highlight the importance of training RL-based signal control models using diverse traffic demand patterns to improve robustness under varying traffic conditions.

This study focuses on a single-intersection control problem. Future work should extend the analysis to multi-agent signal control across multiple intersections, where coordination between intersections introduces additional complexity. In addition, future research should evaluate the robustness of RL-based signal control to imperfect traffic data and sensor failures, which are common in real-world deployments.

## 7  DATA AVAILABILITY STATEMENT

Some or all data, models, or code that support the findings of this study are available from the corresponding author upon reasonable request. This includes:

- Python code for the formulated model
- SUMO model files
- Raw data extracted from SUMO